\documentclass[10pt,journal,compsoc]{IEEEtran}
\usepackage{comment}
\usepackage{balance}
\usepackage{ragged2e}
\usepackage{booktabs}
\usepackage{tabularx}
\usepackage{caption}
\usepackage{graphicx}
\usepackage{longtable}
\usepackage{tikz}
\usetikzlibrary{positioning, arrows.meta, shapes.geometric}
\usepackage{xcolor}
\usepackage{balance}
\usepackage[
    colorlinks=true,
    linkcolor=blue,
    citecolor=blue,
    urlcolor=blue
]{hyperref}

\ifCLASSOPTIONcompsoc
  
  \usepackage[nocompress]{cite}
\else
  \usepackage{cite}
\fi
\ifCLASSINFOpdf
\else
\fi
\begin{document}
%
\title{The End of AI Exponentiation: Fluttering
Inside and Outside AI Bubble}
%
%
%
%

\author{Victor Kebande,~\IEEEmembership{Member,~IEEE.}
\IEEEcompsocitemizethanks{\IEEEcompsocthanksitem Victor. Kebande is with the Department
of Computer Science and Engineering, University of Colorado Denver, Colorado,
CO, 80204, USA.\protect\\
E-mail: victor.kebande@ucdenver.edu
}
\thanks{}}

%
%

\markboth{Preprint. This manuscript has been submitted to IEEE Technology and Society Magazine, 2026}%
{Shell \MakeLowercase{\textit{et al.}}: Bare Demo of IEEEtran.cls for Computer Society Journals}
%



\IEEEtitleabstractindextext{%
\begin{abstract}
The exponentiation of Artificial intelligence (AI) in the recent past  has entered a transformative era that has been driven by the growth in large language models (LLMs), large-scale compute infrastructures, and autonomous reasoning systems. However, the rapid acceleration of AI has increasingly shown technological, societal, economic, ethical and infrastructural challenges associated with peak data limitations, rising computational demands, synthetic data recursion, valuation inflation, and societal instability. The traditional scaling paradigms that have powered the modern AI systems are gradually encountering friction in sustaining continuous exponential growth. This paper views ``the end of AI exponentiation,'' thus exploring how it  flutters inside and outside the bubble, where instability emerges within  the AI ecosystem through compute and data-center races, speculative  investments, and the rat-race toward superintelligence, and outside the  ecosystem through labor disruption, governance concerns, public uncertainty, and geopolitical acceleration surrounding future intelligent systems and infrastructures globally.

\end{abstract}

\begin{IEEEkeywords}
Artificial intelligence, AI exponentiation, agentic AI, 
AI scaling, artificial general intelligence, AI governance
\end{IEEEkeywords}}

\maketitle

\IEEEdisplaynontitleabstractindextext

%
\IEEEpeerreviewmaketitle

\IEEEraisesectionheading{\section{Introduction}\label{sec:introduction}}

%
%
%
%
\IEEEPARstart{I}{n} In recent years, Artificial Intelligence (AI) has  entered an unprecedented phase of technological acceleration that is driven by exponential growth in Large Language Models (LLMs), large-scale compute infrastructures, and advances in machine learning architectures \cite{liang2025ai}. In addition, the emergence of the transformer-based systems, autonomous agents, and reasoning-capable AI models has also significantly transformed the digital ecosystem and this has shown breakthroughs across different sectors  like healthcare, cybersecurity, finance, education, manufacturing, and scientific discovery \cite{ehrlich2025forestgpt, zuo2025large}. It has been observed that from the conversational systems to agentic AI architectures, the intelligent systems are increasingly becoming embedded into societal, industrial, and economic infrastructures, making AI one of the most influential technological forces of the modern era.

However, as AI systems continue to scale rigourously, there is a growing concern that has been observed. It has been seen  that the current trajectory of exponential AI growth may increasingly encounter structural, technological, economic, and societal limitations. This expansion of AI has introduced substantial challenges associated with peak data limitations \cite{kebande2026end}, escalating computational demands \cite{falk2026computation}, energy consumption \cite{mohring2026environmental}, synthetic data recursion, valuation inflation, infrastructure saturation, and governance uncertainty and others \cite{videgaray2024artificial, storm2025us}. Consequently, it has also been seen that, while the frontier AI laboratories continue to pursue increasingly larger and more capable models, there are questions that are emerging regarding the sustainability of scaling-driven AI development and whether the present exponential trajectory can continue indefinitely.

In addition to this, the growing race toward Artificial General Intelligence (AGI), autonomous reasoning systems, and agentic AI has intensified speculative investments, geopolitical competition, and societal uncertainty \cite{amodei2026policy}. As a result,  this has created instability both within the AI ecosystem through compute races, frontier model competition, and economic speculation, and outside the AI ecosystem through labor disruption, educational transformation, regulatory challenges, and public anxiety surrounding the societal impact of intelligent systems \cite{zhang2026ai}. These factors make the current AI landscape to increasingly resemble a transitional phase characterized by technological turbulence and uncertainty.

\begin{figure*}[t]
    \centering
    \includegraphics[width=0.82\textwidth]{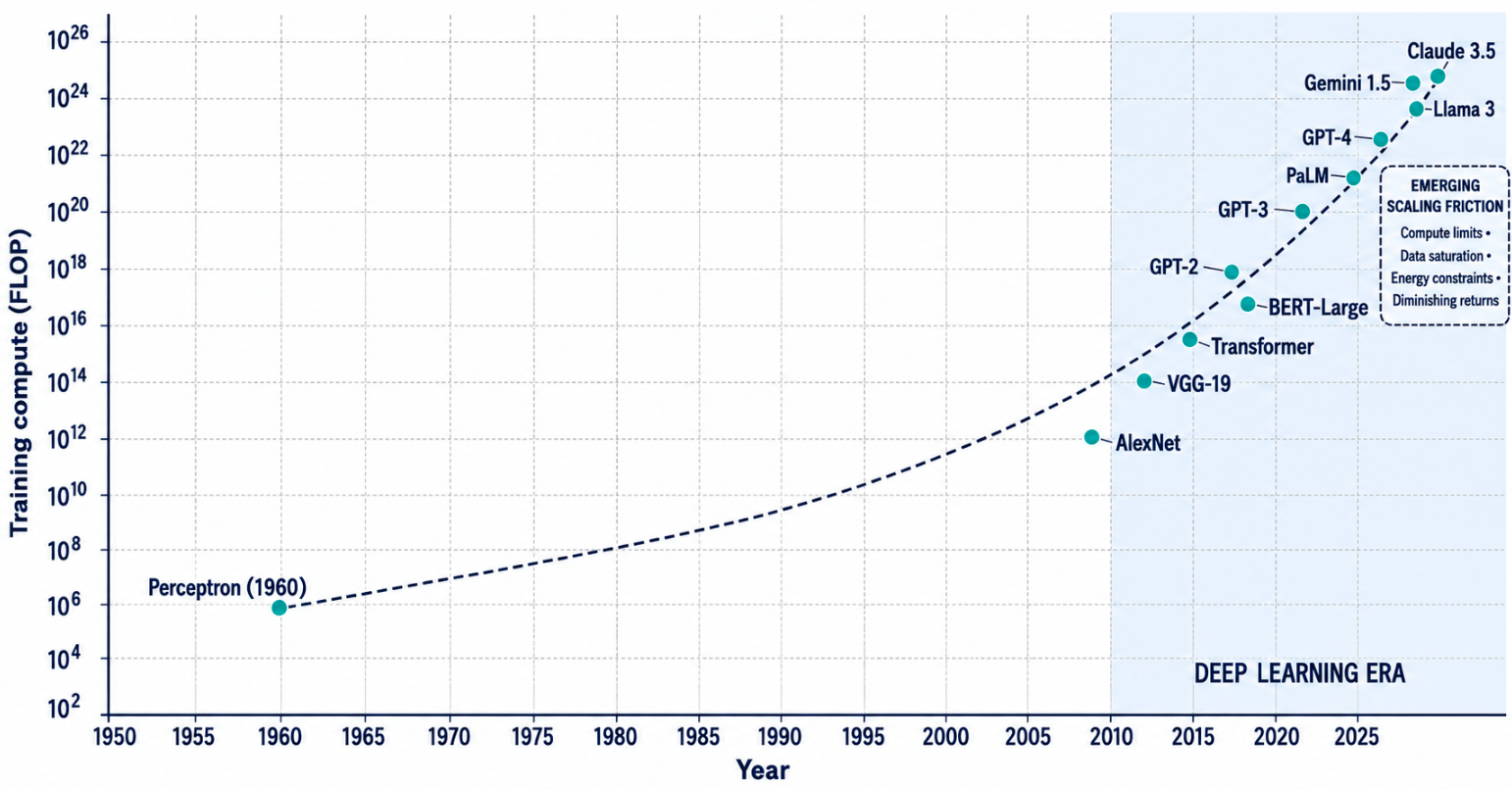}
    \caption{Historical growth of training compute showing the current AI exponential and emerging scaling friction, Epoch AI,  \cite{EpochAIModels2026}.}
    \label{fig:ai_exponential}
\end{figure*}

The goal of this paper, therefore, is to introduce the concept of ``the end of AI exponentiation'' and explore the notion of fluttering inside and outside the bubble. In the context of this paper, fluttering refers to the instability, oscillation, and turbulence emerging across technological, societal, and economic dimensions of AI development. By examining the evolution of scaling-driven AI systems, identifying growing structural limitations, and discussing emerging post-exponential pathways, this paper aims to give the reader a broader perspective on the future trajectory of intelligent systems beyond the  traditional scaling paradigms.

The contribution of this article can be summarised as follows:

\begin{itemize}

\item The author introduces the concept of the ``end of AI exponentiation'' and identifies the emerging limits of scaling-driven AI development.

\item The author describes the concept of  ``fluttering inside and outside the AI bubble'' to characterize technological, economic, and societal instability surrounding AI growth.

\item The author explores post-exponential pathways through agentic AI, reasoning-centric architectures, and adaptive intelligent systems.

\end{itemize}

The remainder of this paper is organized as follows. 
Sections II--V discuss the background, emerging challenges, 
and internal and external pressures surrounding AI exponentiation. 
Sections VI--VIII examine scaling saturation and emerging 
post-exponential AI paradigms. Section IX discusses the ethical, 
economic, and geopolitical implications, while Sections X and XI 
present future directions and conclude the paper, respectively.

\section{Background and Motivation}

\subsection{AI Exponentiation: Importance and Role}

The exponentiation of AI in the recent past has been seen to be  one of the defining technological trends of the modern digital era \cite{makridakis2017forthcoming}. Also, the  rapid advancement of LLMs, autonomous systems, and reasoning-capable architectures has transformed AI from a normal specialized research field into a global technological infrastructure. These infrastructures increasingly influence industries, governments, economies, and societies in diverse ways \cite{chalmers2026acceleration}. In addition, this exponential growth has mostly been driven by advances in transformer architectures, large-scale pretraining, high-performance compute infrastructures, cloud ecosystems, and the availability of Internet-scale datasets. Observations have shown that AI systems are now capable of performing tasks that involve language understanding, software generation, scientific discovery, cybersecurity analysis, autonomous decision-making, and multimodal reasoning \cite{huang2025foundation}.

The role played by AI exponentiation is significant because it has accelerated innovation and productivity across several domains like healthcare, education, finance, manufacturing, transportation, cybersecurity, and critical infrastructure \cite{abbas2024impact}. 

The frontier AI systems seem to be increasingly integrated into industrial and societal ecosystems where intelligent systems support automation, predictive analysis, adaptive reasoning, and human-computer interaction. At the same time, the emergence of agentic AI and reasoning-centric systems has intensified discussions surrounding artificial general intelligence (AGI), autonomous intelligence, and future superintelligent systems \cite{amodei2026policy}.

In spite of this  continued scaling of AI systems, there has been an  increased  number of concerns regarding sustainability, reliability, governance, and long-term feasibility. Owing to the fact that exponential AI growth has generated significant optimism and investment, it has simultaneously created technological and societal instability associated with rising computational demands, energy consumption, synthetic data recursion, infrastructure saturation, and speculative economic expansion.

\begin{table*}[t]
\centering
\caption{Challenges and Friction in AI Exponentiation}
\label{tab:ai_friction}
\begin{tabularx}{\textwidth}{p{3.2cm}X X}
\toprule
\textbf{Challenge} & \textbf{Description} & \textbf{Impact on AI Exponentiation} \\
\midrule
Peak Data \cite{kebande2026end} & High-quality human-generated data is finite and may not continue growing at the same rate as model demand. & Limits the effectiveness of conventional pretraining and may reduce marginal gains from scaling. \\
\midrule
Compute Costs \cite{sastry2024computing} & Frontier AI systems require massive computational infrastructure, specialized chips, and expensive data centers. & Increases barriers to entry and concentrates AI development among a few powerful actors. \\
\midrule
Energy Demand \cite{rozycki2025energy} & AI training and inference consume increasing amounts of electricity and cooling resources. & Raises sustainability concerns and links AI progress to energy infrastructure limits. \\
\midrule
Synthetic Data Recursion \cite{bahov2025model} & Future models may increasingly train on AI-generated content rather than original human-generated data. & May introduce model degradation, bias amplification, and reduced informational diversity. \\
\midrule
Benchmark Saturation \cite{ott2022mapping} & Existing benchmarks may no longer accurately measure progress as models optimize for known evaluation tasks. & Creates uncertainty about whether measured capability reflects real-world intelligence. \\
\midrule
Reasoning Limitations \cite{ke2025survey} & LLMs can generate plausible outputs but may still struggle with deep causal reasoning, long-horizon planning, and robust abstraction. & Suggests that scaling alone may not be enough to achieve reliable autonomous intelligence. \\
\midrule
Economic Speculation \cite{rao2025ai}  & Large investments and high valuations may exceed near-term revenue and deployment realities. & Creates bubble-like instability around AI markets, startups, and infrastructure expansion. \\
\midrule
Governance Lag \cite{bengio2024managing} & Regulation, policy, and institutional adaptation move slower than frontier AI development. & Produces uncertainty around safety, accountability, liability, and responsible deployment. \\
\midrule
Societal Disruption \cite{selgas2025sociotechnical} & AI adoption affects labor markets, education, creativity, security, and public trust. & Expands the consequences of AI exponentiation beyond technical communities. \\
\bottomrule
\end{tabularx}
\end{table*}

\subsection{ Need for Understanding the End of AI Exponentiation}

The rapid acceleration of AI development has increasingly raised questions regarding whether current scaling-driven paradigms can continue sustaining exponential progress indefinitely \cite{genewein2026agi}. Traditional AI growth has largely depended on increasing model sizes, expanding datasets, and scaling computational resources. However, recent discussions within the AI community have suggested that these approaches may gradually encounter structural friction associated with finite high-quality human-generated data, diminishing returns, and escalating infrastructure costs.

At the same time, the growing competition among frontier AI laboratories has intensified compute races, valuation inflation, and speculative investment surrounding AGI and autonomous intelligence systems as is shown in Figure \ref{fig:ai_exponential}. Figure~\ref{fig:ai_exponential} illustrates the exponential growth in training compute, from the early Perceptron to deep-learning models such as AlexNet and Transformer, and frontier systems including GPT-4, Llama 3, Gemini 1.5, and Claude 3.5 \cite{EpochAIModels2026}. The recent trajectory highlights emerging scaling friction associated with compute limitations, data saturation, energy constraints, and diminishing returns. The models are as seen at the time of writing this paper.  These models seem to be entering a scaling or friction phase characterized by compute limits, data saturation, energy constraints, and diminishing returns as is shown from 2025  onwards in Figure~\ref{fig:ai_exponential}. While these developments have accelerated innovation, they have also created instability both inside and outside the AI ecosystem. Inside the AI ecosystem, turbulence emerges through aggressive scaling competition, concentration of compute resources, and uncertainty regarding future capability breakthroughs. Outside the AI ecosystem, instability appears through labor disruption, educational transformation, regulatory uncertainty, public anxiety, and geopolitical competition associated with intelligent systems.

Motivated by these growing concerns, this paper is positioned to explore the concept of possibly ``the end of AI exponentiation'' and then the author analyzes the technological, societal, and economic turbulence emerging across the AI landscape. 


\section{Challenges and Friction in AI Exponentiation}

According to \cite{tu2024overview}, this exponential growth of AI is driven by larger models, greater computational capacity, broader datasets, and improved training techniques \cite{tu2024overview}. In this paper, \textit{AI exponentiation} refers to the rapid compounding of AI capabilities, the investment, the infrastructure, and the societal influence it possesses. This trajectory has and still keep producing remarkable advances in language understanding, code generation like the vibe-coding approaches, multimodal reasoning, and task automation, while fueling expectations surrounding artificial general intelligence (AGI), agentic AI, and reasoning-capable systems. It is nevertheless showing signs of friction that challenge the sustainability of continued exponential progress.

Another challenge is that, the high-quality human-generated data may be approaching saturation, potentially limiting further gains from conventional pretraining as highlighted by \cite{kebande2026end}. This is because, training increasingly capable models  requires massive computational infrastructure, which creates substantial barriers for emerging AI teams \cite{sastry2024computing}, also, the growing energy and cooling demands raise serious sustainability concerns \cite{rozycki2025energy}. The challenge is total reliance of this data.  Greater reliance on AI-generated training data may further degrade model quality over successive generations \cite{bahov2025model}. Moreover, persistent limitations in deep causal reasoning suggest that scaling alone may not produce autonomous intelligence \cite{ott2022mapping}, the rat-race to superintelligence may be in jeopardy. These technical and infrastructural constraints are compounded by speculative investment and inflated valuations that contribute to economic instability \cite{rao2025ai}, as well as regulatory delays that create uncertainty surrounding accountability, security, and responsibility \cite{bengio2024managing}. Their effects extend beyond technical communities, disrupting education, labor, markets, and broader social institutions \cite{selgas2025sociotechnical}.

Table~\ref{tab:ai_friction} summarizes the major sources of friction associated with AI exponentiation while highlighting the respective impacts. Collectively, these pressures suggest that the future of AI may not follow a simple continuation of existing scaling curves. Instead, AI development may be entering a transitional phase in which progress continues under growing technological, economic, infrastructural, and societal constraints.

\begin{figure*}[!t]
    \centering
    \includegraphics[width=\textwidth]{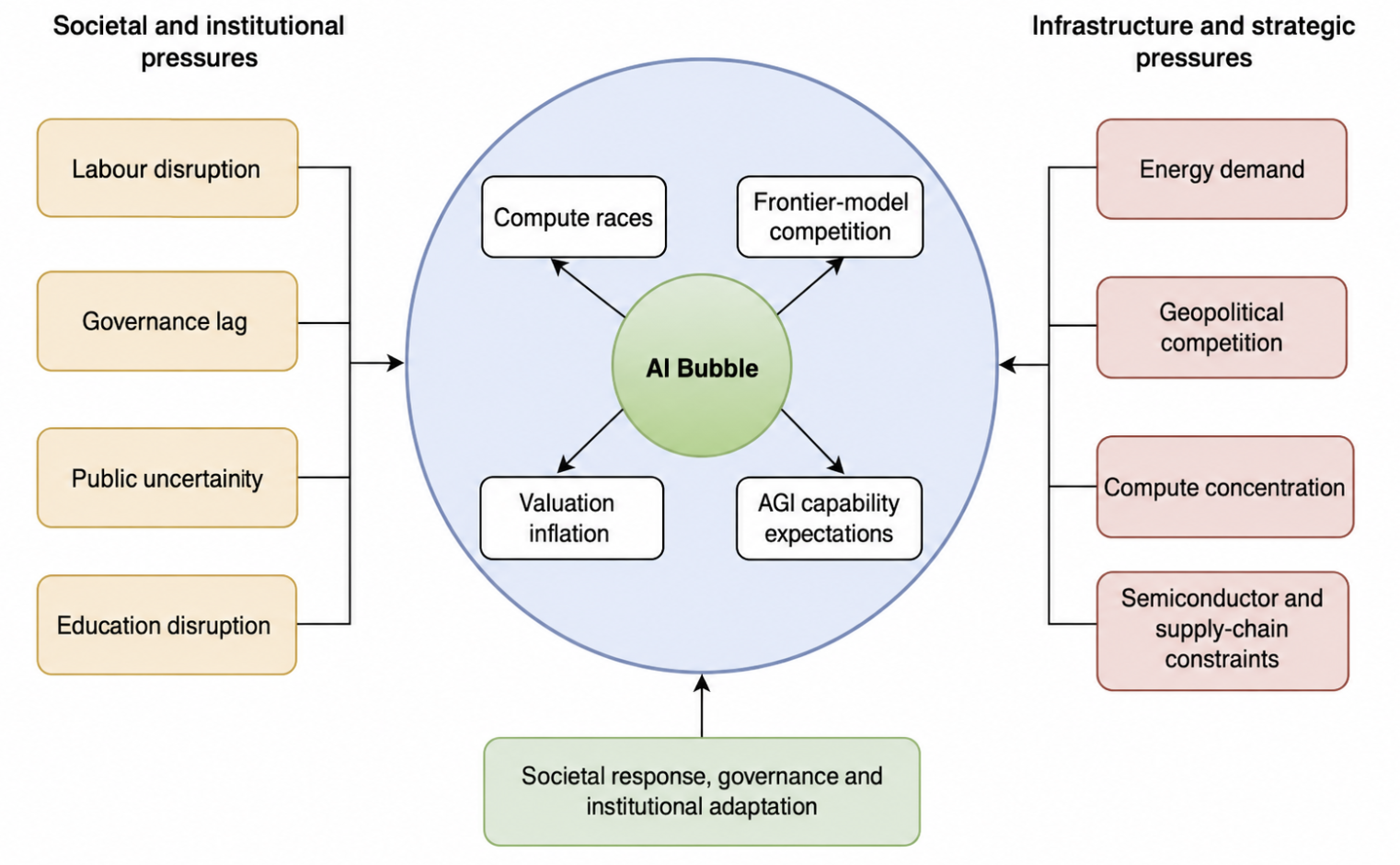}
    \caption{Fluttering inside and outside the AI bubble, showing the internal forces and external pressures shaping AI exponentiation.}
    \label{fig:ai-bubble}
\end{figure*}

\section{Inside the Bubble: Compute Races, Speculation, and Frontier AI Competition}

The internal dynamics of the AI ecosystem are increasingly shaped by competition among frontier AI laboratories, cloud providers, semiconductor manufacturers, investors, and platform companies. This environment constitutes the space ``inside the bubble,'' where expectations surrounding AGI, superintelligence, and autonomous reasoning stimulate aggressive investment, infrastructure expansion, and rapid model deployment \cite{cheng2024appraising}. Figure~\ref{fig:ai-bubble} illustrates how these internal forces interact with broader societal, institutional, infrastructural, and strategic pressures.

Inside the bubble, technological progress is commonly measured through model scale, benchmark performance, computational capacity, and deployment reach. These measures have intensified competition for larger training runs, advanced processors, proprietary datasets, specialized talent, and integrated deployment platforms. Although this competition has accelerated innovation, it has also concentrated frontier AI development among a small number of organizations capable of financing and operating the required infrastructure.

Compute races and frontier-model competition reinforce one another. Each major model release creates pressure on competing laboratories to demonstrate greater capability, efficiency, or market relevance \cite{zhang2026ai}. Cloud providers and semiconductor manufacturers also benefit from this competition because increasingly complex models require extensive computing, storage, networking, and energy resources. However, the dependence on a limited number of cloud platforms and advanced-chip suppliers introduces infrastructural bottlenecks and gives a relatively small group of organizations substantial influence over the direction of AI development.

Valuation inflation represents the economic dimension of the bubble. AI companies are frequently valued not only according to present revenue and demonstrated capabilities, but also according to their anticipated leadership in automation,  scientific discovery, enterprise productivity, and AGI development \cite{grebe2023artificial, moro2024valuation}. This creates tension between technological reality and economic expectation. Continued improvements may justify some investments.


The AGI capability expectations further have amplified this uncertainty. Predictions about imminent autonomous or superintelligent systems have encouraged organizations to accelerate development and secure strategic advantage before competitors \cite{bikkasani2025navigating}. At the same time, there is uncertainty about when or whether such capabilities will emerge and this makes it difficult to distinguish sustainable technological progress from speculative enthusiasm.

\begin{table*}[t]
\centering
\caption{Key dimensions of instability and response inside and outside the AI bubble.}
\label{tab:inside_outside}
\begin{tabularx}{\textwidth}{p{3.2cm}X X}
\toprule
\textbf{Dimension} & \textbf{Inside the Bubble} & \textbf{Outside the Bubble} \\
\midrule
Primary Actors & Frontier AI labs, hyperscalers, chip companies, investors, platform firms. & Workers, students, educators, regulators, governments, citizens, small businesses. \\
\midrule
Main Pressure & Capability acceleration, compute access, model leadership, market dominance. & Adaptation, trust, regulation, employment uncertainty, social stability. \\
\midrule
Instability Form & Speculation, infrastructure race, secrecy, valuation inflation, AGI competition. & Labor disruption, policy confusion, public anxiety, misinformation, dependency. \\
\midrule
Main Risk & Overinvestment, concentration of power, unsafe deployment, unrealistic expectations. & Unequal impact, social displacement, regulatory delay, reduced public trust. \\
\midrule
Possible Response & Responsible scaling, safety evaluation, transparency, energy-aware AI design. & Public literacy, governance frameworks, reskilling, institutional adaptation. \\
\bottomrule
\end{tabularx}
\end{table*}

The concept of \textit{fluttering} captures these internal oscillations as is shown in Table~\ref{tab:inside_outside}. Rather than following a smooth and predictable trajectory, the AI ecosystem moves between optimism and anxiety, openness and secrecy, acceleration and safety, and investment and uncertainty. These competing forces indicate that exponential expectations are increasingly colliding with technical, economic, and infrastructural constraints.

\section{Outside the Bubble: Societal Instability and Technological Disruption}

The consequences of AI exponentiation extend beyond frontier laboratories and technology companies. As Figure~\ref{fig:ai-bubble} shows, pressures outside the bubble can be divided into two interconnected categories: societal and institutional pressures, and infrastructural and strategic pressures. These external forces both respond to developments inside the bubble and influence its future direction.

Societal and institutional pressures include labor disruption, education disruption, governance lag, and public uncertainty. For example, in labor markets, AI creates uncertainty about which tasks will be automated, which occupations will be reorganized, and which skills will retain long-term value \cite{frank2019toward}. Although AI may create new forms of employment and improve productivity, its benefits and costs may be distributed unevenly across industries, professions, and communities.

In education, generative AI challenges established assumptions about authorship, assessment, learning, and academic integrity \cite{kofinas2025impact}. Educational institutions must reconsider how student knowledge is evaluated and how AI tools should be incorporated into teaching and research. Without appropriate adaptation, the reliance on AI-generated content may weaken independent reasoning, or may create uncertainty about intellectual ownership, and widen inequalities between students with different levels of access and AI literacy.

Governance systems face a similar challenge. Policymakers are attempting to regulate AI systems whose capabilities, architectures, and applications evolve more rapidly than conventional legislative and regulatory processes \cite{gaske2023regulation}. This governance lag creates uncertainty concerning accountability, transparency, privacy, security, intellectual property, and responsibility for AI-assisted decisions. It may also allow potentially harmful systems to be deployed before appropriate evaluation and oversight mechanisms are established.

Public uncertainty emerges because many individuals interact with AI systems without fully understanding how they operate, how their outputs are produced, or where their limitations lie. This uncertainty can generate fear, exaggerated expectations, misplaced trust, and growing dependency. It may also make it difficult for the public to distinguish demonstrated capabilities from speculative claims concerning AGI and autonomous intelligence.

The infrastructural and strategic pressures outside the bubble include energy demand, geopolitical competition, compute concentration, and semiconductor supply-chain constraints \cite{jeppesenpolitical}. Training and operating frontier models require substantial electricity, cooling, data-center capacity, and specialized hardware. As AI adoption expands, these requirements raise questions about environmental sustainability, infrastructure availability, and competition with other demands for energy and water.

AI has also become an area of geopolitical competition. Governments increasingly view advanced models, semiconductor production, cloud infrastructure, and technical expertise as strategically important national assets \cite{horowitz2022strategic}. Export controls, investment restrictions, and national AI programs demonstrate how competition for AI leadership is reshaping international relations. At the same time, the concentration of computing resources among a few countries and corporations may deepen global inequalities in access to advanced AI capabilities.

Semiconductor and supply-chain constraints further expose the physical limits underlying seemingly digital AI systems \cite{wyon2025new, rottensteiner2025great}. Frontier development depends on specialized chips, manufacturing equipment, critical materials, global logistics, and geographically concentrated production facilities. Disruption at any point in this chain can affect model development, infrastructure expansion, and access to computational resources.

Cybersecurity connects the internal and external dimensions of the bubble. AI provides new capabilities for threat detection, vulnerability analysis, incident response, and defensive automation. However, it can also support reconnaissance, social engineering, phishing, vulnerability discovery, malware development, and automated exploitation. The integration of autonomous agents into security-sensitive environments may therefore increase both defensive capacity and systemic risk.

\begin{figure*}[!t]
    \centering
    \includegraphics[width=\textwidth]{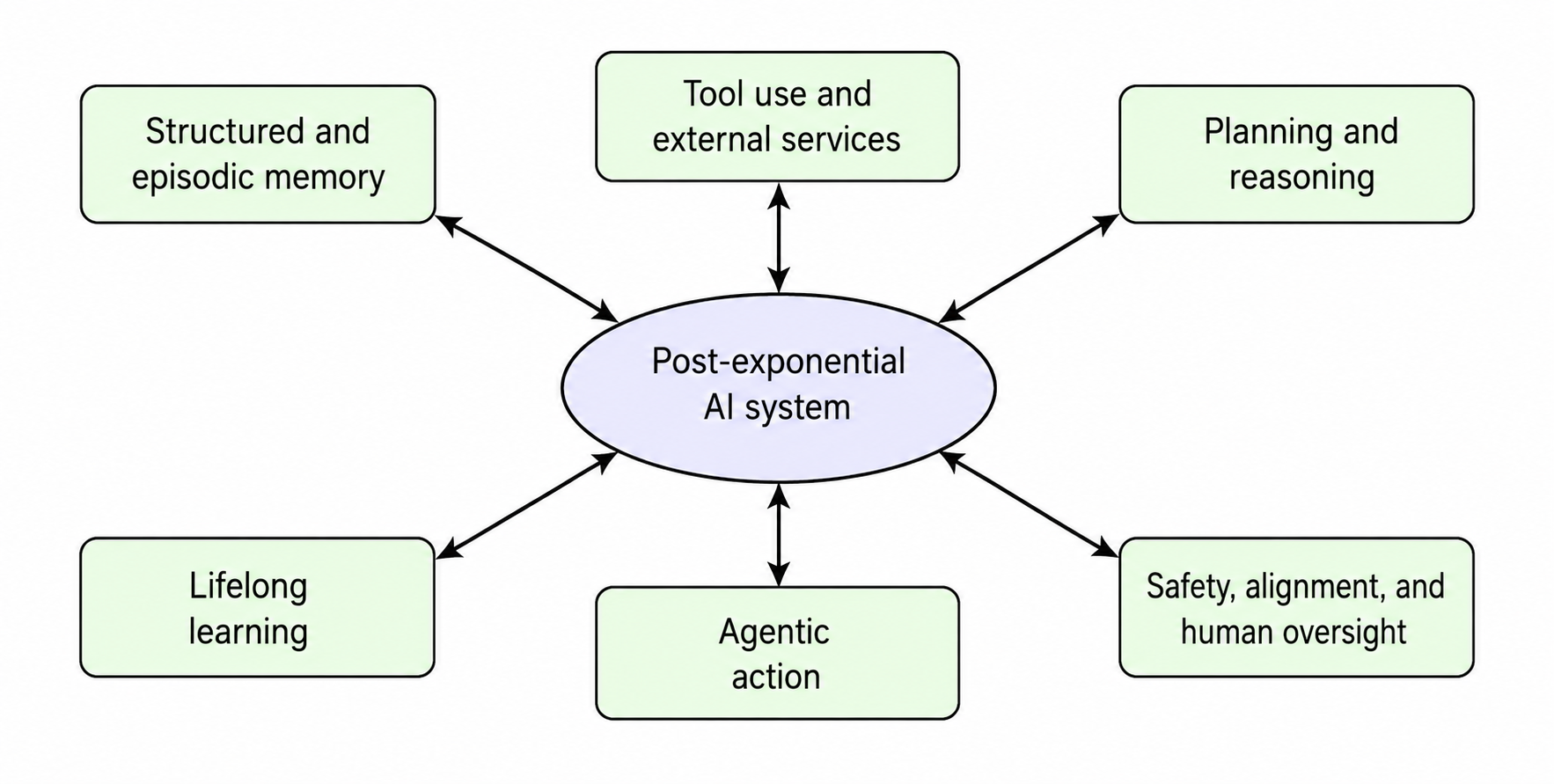}
   \caption{Conceptual architecture for post-exponential AI integrating memory, reasoning, tool use, learning, agentic action, and safety.}
    \label{fig:post-exponential-ai-system}
\end{figure*}

Societal responses, governance mechanisms, and institutional adaptation create a feedback relationship between the outside and inside of the bubble. Regulation can influence model development and deployment, public resistance can affect adoption, energy limitations can constrain infrastructure expansion, and educational or workforce policies can shape how societies respond to automation. The boundary of the bubble is therefore permeable: internal developments generate external consequences, while external pressures reshape internal priorities and practices.

Table~\ref{tab:inside_outside} contrasts the principal actors, pressures, risks, and possible responses associated with the environments inside and outside the AI bubble. The distinction between these environments is important because AI exponentiation does not occur in isolation. Its momentum is generated within concentrated technical and financial systems, while its consequences are distributed across society, institutions, infrastructure, and international relations. Any account of the end of AI exponentiation must therefore consider both the internal constraints affecting AI development and the external disruptions arising from its deployment.

\section{Peak Data, Synthetic Recursion, and Scaling Saturation}

One of the most important sources of friction in AI exponentiation is the possibility of peak data. Modern LLMs have relied heavily on large-scale pretraining using human-generated text, code, images, and multimodal content. However, the supply of high-quality human-generated data is finite, but \cite{kebande2026end} has argued that pretraining may end as a result of peak data. As models become larger and training demands increase, the availability of clean, diverse, reliable, and legally usable data seen as the fossil fuel for AI becomes a major constraint.

This challenge is closely connected to synthetic data recursion. As the Internet becomes increasingly populated with AI-generated content, future models may train on outputs generated by earlier models. While synthetic data can be useful for augmentation, simulation, and specialized reasoning tasks, excessive dependence on synthetic content may weaken model diversity and increase the risk of error reinforcement. If models repeatedly learn from model-generated outputs, they may amplify biases, hallucinations, stylistic uniformity, and factual distortions.

Scaling saturation also appears in evaluation. Many benchmarks that once measured meaningful progress are now saturated by frontier models. This creates difficulty in distinguishing between genuine reasoning ability and benchmark-specific optimization. A model may perform well on standardized tests but still fail in open-ended, noisy, ambiguous, or long-horizon real-world environments.

Figure~\ref{fig:post-exponential-ai-system} presents a simplified view of the transition from scaling-driven AI toward post-exponential AI systems. Rather than relying primarily on larger models, datasets, and computational resources, this architecture emphasizes the integration of memory, tool use, planning and reasoning, lifelong learning, and agentic action. These capabilities interact to support more adaptive and autonomous AI behavior, while safety, alignment, and human oversight provide mechanisms for governing increasingly capable systems.

Thus, the end of AI exponentiation should  not be interpreted as the end of AI progress. Rather, it may represent the end of a familiar scaling regime. Future progress may depend less on simply increasing model size and more on developing architectures that can reason, interact, remember, plan, verify, and adapt.

\begin{figure*}[!t]
    \centering
    \includegraphics[width=0.98\textwidth]{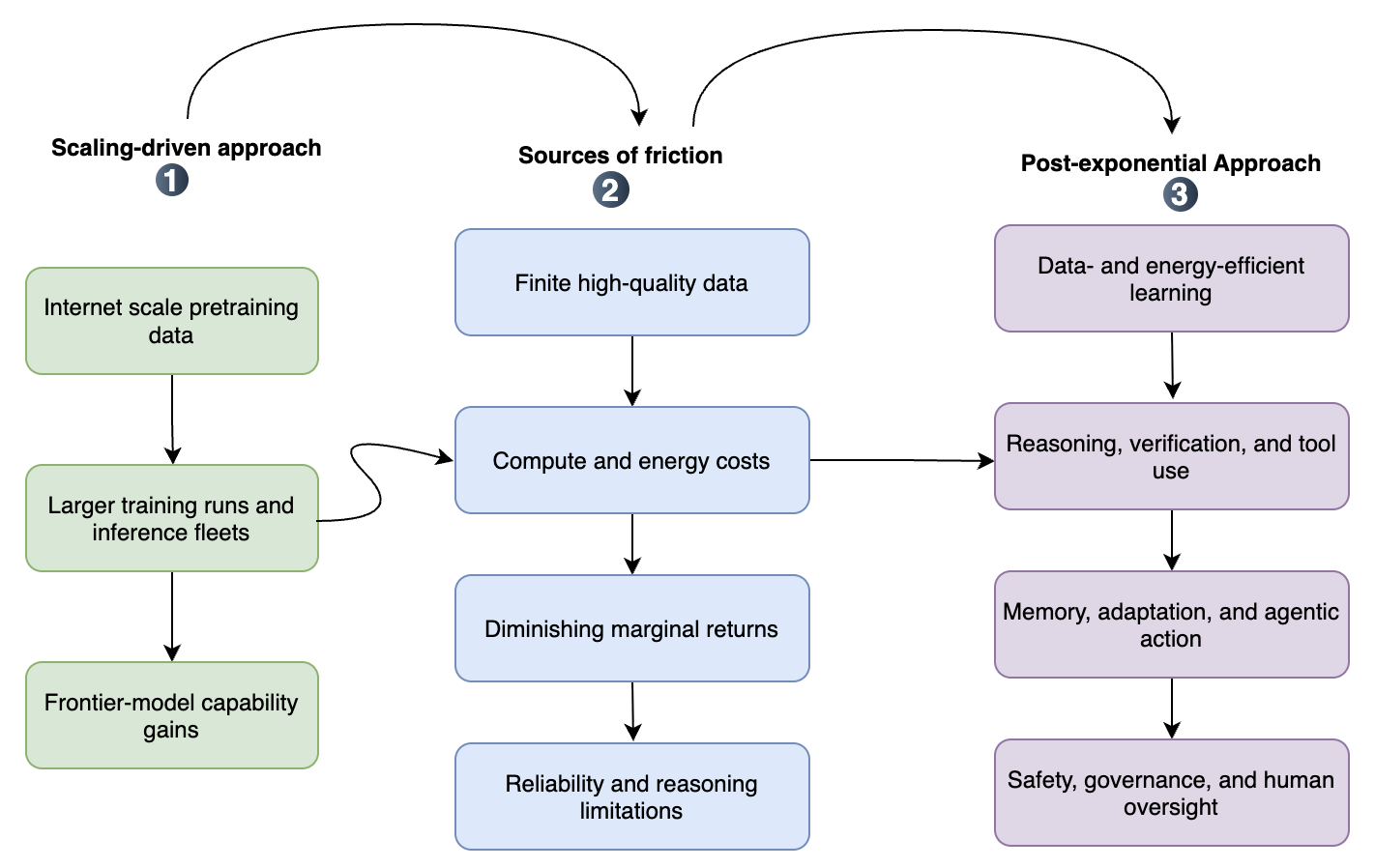}
    \caption{Transition from scaling-driven AI toward efficient, adaptive, and reasoning-centric post-exponential AI systems.}
    \label{fig:post_exponential_transition}
\end{figure*}

\section{Beyond Exponentiation: Agentic AI and Reasoning-Centric Architectures}

As scaling-driven progress encounters friction, attention is shifting toward agentic AI and reasoning-centric architectures. Agentic AI refers to systems that can pursue goals, use tools, interact with environments, make decisions, and adapt their actions based on feedback. Unlike conventional prompt-response systems, agentic systems are designed to operate across sequences of tasks, often with memory, planning, and tool-use capabilities.

Reasoning-centric architectures aim to move AI beyond statistical pattern recognition. Although LLMs can produce fluent and useful responses, they may still struggle with causal reasoning, formal verification, long-term planning, and consistent decision-making under uncertainty. Reasoning-centric systems attempt to address these limitations through methods such as tool use, structured memory, symbolic reasoning, reinforcement learning, causal models, and human-in-the-loop feedback.

This transition is illustrated in Figure~\ref{fig:post_exponential_transition}, 
which shows the shift from scaling-driven AI (labeled 1 to 3), through emerging sources of 
friction, toward more efficient, adaptive, and reasoning-centric 
post-exponential systems.

The relevance of this direction is that it reframes the future of AI. The question is no longer only whether models can become larger, but whether AI systems can become more reliable, autonomous, interpretable, aligned, and useful across real-world environments.

\section{Autonomy, Adaptive Intelligence, and Post-Scaling AI Systems}

Post-scaling AI systems are likely to be defined by autonomy and adaptation. Autonomy allows AI systems to initiate actions, decompose tasks, use tools, and pursue goals with reduced human intervention. Adaptive intelligence allows systems to update their behavior based on feedback, context, and environmental changes.

This shift has important implications. In the scaling era, AI progress was often associated with training larger models on larger datasets. In the post-scaling era, progress may depend on how well systems can operate in dynamic environments. Such systems may need persistent memory, contextual awareness, verification loops, multimodal perception, and the ability to collaborate with humans and other agents.

However, autonomy also introduces risk. The more capable and independent AI systems become, the more important it becomes to ensure that their goals remain aligned with human values. Autonomous systems may behave unpredictably if their objectives are poorly specified, if they operate in unfamiliar environments, or if they optimize for narrow goals without understanding broader consequences.

Therefore, the movement beyond exponentiation requires not only technical innovation but also careful governance. Post-scaling AI must be designed with transparency, controllability, accountability, and safety as core requirements. Without these safeguards, autonomous systems may amplify the same instability that characterizes the current AI bubble.

\begin{table*}[t]
\centering
\caption{Ethical, Economic, and Geopolitical Implications of the End of AI Exponentiation}
\label{tab:implications}
\begin{tabularx}{\textwidth}{p{3cm}X X}
\toprule
\textbf{Implication Area} & \textbf{Key Concern} & \textbf{Possible Response} \\
\midrule
Ethics & Bias, opacity, misalignment, privacy loss, and reduced human agency. & Transparent evaluation, human oversight, alignment research, and explainable AI. \\
\midrule
Economics & Market speculation, job displacement, concentration of wealth, and unequal access. & Reskilling programs, fair access policies, productivity-sharing models, and responsible investment. \\
\midrule
Infrastructure & Energy demand, chip scarcity, data center expansion, and environmental pressure. & Energy-aware AI, efficient models, distributed compute, and sustainable infrastructure planning. \\
\midrule
Security & AI-enabled cyber offense, autonomous reconnaissance, misinformation, and dual-use risks. & AI security standards, red-teaming, monitoring, and defensive AI systems. \\
\midrule
Geopolitics & Strategic competition, military use, export controls, and global governance fragmentation. & International norms, AI treaties, safety cooperation, and cross-border governance mechanisms. \\
\bottomrule
\end{tabularx}
\end{table*}

\section{Ethical, Economic, and Geopolitical Implications}

The end of AI exponentiation carries ethical, economic, and geopolitical implications. Ethically, advanced AI systems raise concerns related to bias, transparency, accountability, human agency, privacy, and safety. If AI systems become more agentic and autonomous, these concerns become more urgent because decisions may be made or influenced by systems whose internal reasoning is difficult to inspect.

From an economic perspective,  the AI bubble reflects both genuine technological promise and speculative uncertainty. AI may produce productivity gains, scientific breakthroughs, and new industries. However, it may also displace workers, concentrate wealth, and intensify inequality if access to advanced systems is controlled by a small number of organizations. The economic impact of AI will therefore depend not only on capability growth but also on distribution, governance, education, and institutional adaptation.

Geopolitically, AI exponentiation has become part of strategic competition among nations. Compute infrastructure, semiconductor supply chains, data governance, military applications, and AI safety standards are increasingly connected to national power. This creates pressure to accelerate deployment even when safety and governance frameworks remain incomplete.

Table~\ref{tab:implications} summarizes the major implication areas. These implications show that the end of AI exponentiation is not only a technical transition. It is a broader societal transition in which intelligent systems become deeply embedded in economic, political, and human systems.

\section{Future Directions}

Several future directions can shape the post-exponential AI landscape. First, there is a need for reasoning-centric AI systems that can go beyond next-token prediction and support causal inference, abstraction, verification, and long-horizon planning. Such systems may combine neural models with symbolic reasoning, external tools, memory systems, and formal verification methods.

Second, future AI research should focus on sustainable scaling. This includes energy-efficient architectures, smaller specialized models, improved inference efficiency, and data-efficient learning. If compute and energy become major constraints, then progress will depend on efficiency rather than size alone.

Third, agentic AI systems require stronger safety mechanisms. Future research should focus on controllability, alignment, interpretability, and monitoring of autonomous behavior. As AI systems become more capable of taking actions in digital and physical environments, safety must be treated as a core design principle rather than an afterthought.

Fourth, society needs stronger governance and adaptation frameworks. This includes AI literacy, workforce reskilling, educational reform, regulatory coordination, and institutional readiness. The future of AI will not be determined only by laboratories and companies, but also by how societies absorb and govern intelligent systems.

Finally, the post-exponential era may require a new way of measuring progress. Instead of focusing only on model size, benchmark scores, or parameter counts, future evaluation should consider reliability, reasoning depth, energy efficiency, safety, social impact, and human benefit.

\section{Conclusion}

This paper has explored the concept of the end of AI exponentiation and introduced the notion of fluttering inside and outside the bubble. The central argument is not that AI progress is ending, but that the familiar scaling-driven regime is encountering increasing friction. Peak data, compute costs, energy demand, synthetic data recursion, reasoning limitations, speculative investment, and societal instability all suggest that AI is entering a more uncertain transitional phase.

Inside the bubble, frontier AI development is shaped by compute races, model competition, valuation inflation, and AGI expectations. Outside the bubble, societies face labor disruption, governance challenges, public uncertainty, educational transformation, and geopolitical pressure. These internal and external dynamics create a form of fluttering that reflects instability during a major technological transition.

The future of AI may therefore depend less on continuous exponentiation and more on the development of sustainable, adaptive, agentic, and reasoning-centric systems. The end of AI exponentiation should not be viewed as a collapse of AI progress, but as the beginning of a new phase in which intelligence becomes more autonomous, more embedded, and more socially consequential. The challenge ahead is to ensure that this transition is technically robust, economically responsible, ethically aligned, and beneficial to society.

\ifCLASSOPTIONcompsoc
  \section*{Acknowledgments}
\else
  \section*{Acknowledgment}
\fi

The author would like to thank anonymous reviewers for their valuable insights, and the Department of Computer Science at University of Colorado Denver, USA for their support while coming up with this research. The opinions, findings, and conclusions expressed in this article are solely those of the author.

\ifCLASSOPTIONcaptionsoff
  \newpage
\fi



%
\balance
\bibliographystyle{IEEEtran}
\bibliography{references}

@article{liang2025ai,
  title={AI Compute Architecture and Evolution Trends},
  author={Liang, Bor-Sung},
  journal={arXiv preprint arXiv:2508.21394},
  year={2025}
}

@article{zuo2025large,
  title={Large language models for batteries},
  author={Zuo, Wenhua and Zheng, Huihuo and He, Tanjin and Vishwanath, Venkatram and Chan, Maria KY and Stevens, Rick L and Amine, Khalil and Xu, Gui-Liang},
  journal={Joule},
  volume={9},
  number={8},
  year={2025},
  publisher={Elsevier}
}

@article{sastry2024computing,
  title={Computing power and the governance of artificial intelligence},
  author={Sastry, Girish and Heim, Lennart and Belfield, Haydn and Anderljung, Markus and Brundage, Miles and Hazell, Julian and O'keefe, Cullen and Hadfield, Gillian K and Ngo, Richard and Pilz, Konstantin and others},
  journal={arXiv preprint arXiv:2402.08797},
  year={2024}
}

@inproceedings{bahov2025model,
  title={Model collapse in the age of synthetic data: Risks and consequences},
  author={Bahov, Bozhidar},
  booktitle={Selected Papers from the 14th International Conference on Application of Information and Communication Technology and Statistics in Economy and Education (ICAICTSEE-2024), December 2-3rd, 2024, UNWE, Sofia, Bulgaria},
  pages={248--256},
  year={2025},
  organization={University for National and Secular Studies (UNSS)}
}

@article{ke2025survey,
  title={A survey of frontiers in llm reasoning: Inference scaling, learning to reason, and agentic systems},
  author={Ke, Zixuan and Jiao, Fangkai and Ming, Yifei and Nguyen, Xuan-Phi and Xu, Austin and Long, Do Xuan and Li, Minzhi and Qin, Chengwei and Wang, Peifeng and Savarese, Silvio and others},
  journal={arXiv preprint arXiv:2504.09037},
  year={2025}
}

@article{bengio2024managing,
  title={Managing extreme AI risks amid rapid progress},
  author={Bengio, Yoshua and Hinton, Geoffrey and Yao, Andrew and Song, Dawn and Abbeel, Pieter and Darrell, Trevor and Harari, Yuval Noah and Zhang, Ya-Qin and Xue, Lan and Shalev-Shwartz, Shai and others},
  journal={Science},
  volume={384},
  number={6698},
  pages={842--845},
  year={2024},
  publisher={American Association for the Advancement of Science}
}

@misc{EpochAIModels2026,
  author       = {{Epoch AI}},
  title        = {{Data on AI Models}},
  year         = {2026},
  month        = sep,
  howpublished = {\url{https://epoch.ai/data/ai-models}},
  note         = {Accessed: September 3, 2026}
}

@article{selgas2025sociotechnical,
  title={Sociotechnical transformation: a systematic review on the impact of artificial intelligence on society and organizations},
  author={Selgas-Cors, Marc},
  journal={FinTech and Sustainable Innovation},
  pages={1--16},
  year={2025}
}

@article{huang2025foundation,
  title={Foundation models and intelligent decision-making: Progress, challenges, and perspectives},
  author={Huang, Jincai and Xu, Yongjun and Wang, Qi and Wang, Qi Cheems and Liang, Xingxing and Wang, Fei and Zhang, Zhao and Wei, Wei and Zhang, Boxuan and Huang, Libo and others},
  journal={The Innovation},
  volume={6},
  number={6},
  year={2025},
  publisher={Elsevier}
}

@article{zhang2026ai,
  title={What the AI Race Has Given Us and What It Requires Next},
  author={Zhang, Xufeng},
  journal={International Journal of Human--Computer Interaction},
  pages={1--22},
  year={2026},
  publisher={Taylor \& Francis}
}

@article{rao2025ai,
  title={Is AI a Bubble That Is About to Burst? A Systematic Review of Financial and Economic Evidence},
  author={Rao, Sohail},
  journal={INNOVAPATH},
  volume={1},
  number={4},
  pages={10--10},
  year={2025}
}

@article{ott2022mapping,
  title={Mapping global dynamics of benchmark creation and saturation in artificial intelligence},
  author={Ott, Simon and Barbosa-Silva, Adriano and Blagec, Kathrin and Brauner, Jan and Samwald, Matthias},
  journal={Nature Communications},
  volume={13},
  number={1},
  pages={6793},
  year={2022},
  publisher={Nature Publishing Group UK London}
}

@article{rozycki2025energy,
  title={Energy-aware machine learning models—a review of recent techniques and perspectives},
  author={R{\'o}{\.z}ycki, Rafa{\l} and Solarska, Dorota Agnieszka and Walig{\'o}ra, Grzegorz},
  journal={Energies},
  volume={18},
  number={11},
  pages={2810},
  year={2025},
  publisher={MDPI}
}

@article{ehrlich2025forestgpt,
  title={Forestgpt and beyond: A trustworthy domain-specific large language model paving the way to forestry 5.0},
  author={Ehrlich-Sommer, Florian and Eberhard, Benno and Holzinger, Andreas},
  journal={Electronics},
  volume={14},
  number={18},
  pages={3583},
  year={2025},
  publisher={MDPI}
}

@article{genewein2026agi,
  title={From agi to asi},
  author={Genewein, Tim and Franklin, Matija and Lerchner, Alexander and Orseau, Laurent and Albanie, Samuel and Bales, Adam and Wyeth, Cole and Chan, Stephanie and Gabriel, Iason and Leibo, Joel Z and others},
  journal={arXiv preprint arXiv:2606.12683},
  year={2026}
}

@incollection{abbas2024impact,
  title={Impact of artificial intelligence on the global economy and technology advancements},
  author={Abbas Khan, Muhammad and Khan, Habib and Omer, Muhammad Faizan and Ullah, Inam and Yasir, Muhammad},
  booktitle={Artificial general intelligence (AGI) security: Smart applications and sustainable technologies},
  pages={147--180},
  year={2024},
  publisher={Springer}
}

@article{amodei2026policy,
  title={Policy on the AI Exponential},
  author={Amodei, Dario},
  journal={darioamodei. com, June},
  year={2026}
}

@article{chalmers2026acceleration,
  title={The acceleration of artificial intelligence: Rethinking organization and work in an era of rapid technological change},
  author={Chalmers, Dominic and Hunt, Richard  and Pachidi, Stella and Poto{\v{c}}nik, Kristina and Townsend, David},
  journal={Journal of Management Studies},
  volume={63},
  number={2},
  pages={285--314},
  year={2026},
  publisher={Wiley Online Library}
}

@article{makridakis2017forthcoming,
  title={The forthcoming Artificial Intelligence (AI) revolution: Its impact on society and firms},
  author={Makridakis, Spyros},
  journal={Futures},
  volume={90},
  pages={46--60},
  year={2017},
  publisher={Elsevier}
}

@article{videgaray2024artificial,
  title={Artificial intelligence and economic and financial policy making},
  author={Videgaray, Luis and Aghion, Philip and Caputo, Barbara and Forrest, Tracey and Korinek, Anton and Langenbucher, Katja and Miyamoto, Hiroaki and Wooldridge, Michael},
  journal={A High-Level Panel of Experts’ Report to the G},
  volume={7},
  pages={G7Italia},
  year={2024}
}

@article{mohring2026environmental,
  title={The environmental impact of AI: a framework for energy consumption in machine learning services},
  author={M{\"o}hring, Michael and Keller, Barbara and Di Pietro, Laura},
  journal={International Journal of Quality and Service Sciences},
  pages={1--23},
  year={2026},
  publisher={Emerald Publishing Limited}
}

@article{falk2026computation,
  title={From computation to environmental cost the resource burden of artificial intelligence},
  author={Falk, Sophia and Kluge Corr{\^e}a, Nicholas and Luccioni, Sasha and Biber-Freudenberger, Lisa and van Wynsberghe, Aimee},
  journal={Communications Earth \& Environment},
  volume={7},
  number={1},
  pages={397},
  year={2026},
  publisher={Nature Publishing Group UK London}
}

@article{kofinas2025impact,
  title={The impact of generative AI on academic integrity of authentic assessments within a higher education context},
  author={Kofinas, Alexander K and Tsay, Crystal Han-Huei and Pike, David},
  journal={British Journal of Educational Technology},
  volume={56},
  number={6},
  pages={2522--2549},
  year={2025},
  publisher={Wiley Online Library}
}

@article{wyon2025new,
  title={New challenges for the semiconductor ecosystem},
  author={Wyon, Christophe and Van de Voorde, Marcel},
  journal={IEEE Electron Devices Reviews},
  year={2025},
  publisher={IEEE}
}

@phdthesis{rottensteiner2025great,
  title={" With Great Power There Must Also Come Great Responsibility": How the Semiconductor Industry Deals with Uncertainty in Light of Geopolitical and Supply Chain Disruptions},
  author={Rottensteiner, Valentin},
  year={2025},
  school={University of Duisburg-Essen Duisburg}
}

@book{horowitz2022strategic,
  title={Strategic competition in an era of artificial intelligence},
  author={Horowitz, Michael C and Allen, Gregory C and Kania, Elsa B and Scharre, Paul},
  year={2022},
  publisher={Center for a New American Security}
}

@article{jeppesenpolitical,
  title={The Political Economy of Artificial Intelligence},
  author={Jeppesen, Jakob Werner and Kirk, Victor Normann},
  year={2025},
}

@article{gaske2023regulation,
  title={Regulation priorities for artificial intelligence foundation models},
  author={Gaske, Matthew R},
  journal={Vand. J. Ent. \& Tech. L.},
  volume={26},
  pages={1},
  year={2023},
  publisher={HeinOnline}
}

@article{grebe2023artificial,
  title={Artificial intelligence: how leading companies define use cases, scale-up utilization, and realize value},
  author={Grebe, Michael and Franke, Marc Roman and Heinzl, Armin},
  journal={Informatik Spektrum},
  volume={46},
  number={4},
  pages={197--209},
  year={2023},
  publisher={Springer}
}

@article{frank2019toward,
  title={Toward understanding the impact of artificial intelligence on labor},
  author={Frank, Morgan R and Autor, David and Bessen, James E and Brynjolfsson, Erik and Cebrian, Manuel and Deming, David J and Feldman, Maryann and Groh, Matthew and Lobo, Jos{\'e} and Moro, Esteban and others},
  journal={Proceedings of the National Academy of Sciences},
  volume={116},
  number={14},
  pages={6531--6539},
  year={2019},
  publisher={National Academy of Sciences}
}

@incollection{moro2024valuation,
  title={The valuation of artificial intelligence},
  author={Moro-Visconti, Roberto},
  booktitle={Artificial intelligence valuation: The impact on automation, BioTech, ChatBots, FinTech, B2B2C, and other industries},
  pages={405--506},
  year={2024},
  publisher={Springer}
}

@article{cheng2024appraising,
  title={Appraising regulatory framework towards artificial general intelligence (AGI) under digital humanism},
  author={Cheng, Le and Gong, Xuan},
  journal={International Journal of Digital Law and Governance},
  volume={1},
  number={2},
  pages={269--312},
  year={2024},
  publisher={De Gruyter}
}

@article{bikkasani2025navigating,
  title={Navigating artificial general intelligence (AGI): Societal implications, ethical considerations, and governance strategies},
  author={Bikkasani, Dileesh Chandra},
  journal={AI and Ethics},
  volume={5},
  number={3},
  pages={2021--2036},
  year={2025},
  publisher={Springer}
}

@article{kebande2026end,
  title={The End of Pretraining for Large Language Models: The Future of Agentic and AI Reasoning Beyond Peak Data},
  author={Kebande, Victor},
  journal={Computer},
  volume={59},
  number={3},
  pages={60--69},
  year={2026},
  publisher={IEEE}
}

@article{storm2025us,
  title={The US is betting the economy on ‘Scaling’AI: where is the intelligence when one needs it?},
  author={Storm, Servaas},
  journal={International Journal of Political Economy},
  volume={54},
  number={4},
  pages={425--452},
  year={2025},
  publisher={Taylor \& Francis}
}

@article{tu2024overview,
  title={An overview of large AI models and their applications},
  author={Tu, Xiaoguang and He, Zhi and Huang, Yi and Zhang, Zhi-Hao and Yang, Ming and Zhao, Jian},
  journal={Visual Intelligence},
  volume={2},
  number={1},
  pages={34},
  year={2024},
  publisher={Springer}
}

%

\begin{IEEEbiography}{VICTOR KEBANDE}
is a cybersecurity researcher and assistant professor of Cybersecurity at University of Colorado Denver, Denver Colorado, USA, ATLAS Institute, University of Colorado Boulder, Boulder, CO, USA; and University of Colorado Denver, Denver, CO, USA. His research interests include cybersecurity, digital forensics in the Internet of Things, artificial intelligence in cybersecurity, critical infrastructure protection,and cloud security. Kebande received a Ph.D. in computer science (informationand computer security architectures and digital forensics) in 2018. He serves on the Editor of Forensic Science International: Reports. He is a Member of IEEE. Contact him at victor.kebande@ucdenver.edu, victor.kebande@colorado.edu
\end{IEEEbiography}

\vspace{1em}

\noindent
\raisebox{0.1ex}{\rule{1.2ex}{1.2ex}}\hspace{0.5em}
Direct questions and comments about this article to Victor Kebande,
University of Colorado Denver, Denver, CO, USA;
\href{Email:}{victor.kebande@ucdenver.edu}.






\end{document}